\documentclass[conference]{IEEEtran}
\IEEEoverridecommandlockouts
\usepackage{cite}
\usepackage{amsmath,amssymb,amsfonts}
\usepackage{algorithmic}
\usepackage{graphicx}
\usepackage{textcomp}
\usepackage{xcolor}
\def\BibTeX{{\rm B\kern-.05em{\sc i\kern-.025em b}\kern-.08em
    T\kern-.1667em\lower.7ex\hbox{E}\kern-.125emX}}
\begin{document}

\title{Conversational Norms for Human-Robot Dialogues\\
\thanks{This work has received funding from the European Union's
  Horizon 2020 research and innovation programme under the 
Marie Sk\l{}odowska-Curie grant agreement No 721619 for the SOCRATES project.}
}

\author{\IEEEauthorblockN{1\textsuperscript{st} Maitreyee Tewari}
\IEEEauthorblockA{\textit{Department of Computing Science} \\
\textit{Ume{\aa} University}\\
Ume{\aa}, Sweden\\
maittewa@cs.umu.se}
\and
\IEEEauthorblockN{2\textsuperscript{nd} Thomas Hellstr\"om}
\IEEEauthorblockA{\textit{Department of Computing Science} \\
\textit{Ume{\aa} University}\\
Ume{\aa}, Sweden \\
thomash@cs.umu.se}
\and
\IEEEauthorblockN{3\textsuperscript{rd} Suna Bensch}
\IEEEauthorblockA{\textit{Department of Computing Science} \\
\textit{Ume{\aa} University}\\
Ume{\aa}, Sweden\\
suna@cs.umu.se}
}

\maketitle

\begin{abstract}
This paper describes a recently initiated research project aiming at supporting development of computerised dialogue
systems that handle breaches of conversational norms such
as the Gricean maxims, which describe how dialogue
participants ideally form their utterances in order to be infor-
mative, relevant, brief, etc. Our approach is to model dialogue
and norms with co-operating distributed grammar systems
(CDGSs), and to develop methods to detect breaches and to
handle them in dialogue systems for verbal human-robot
interaction.
\end{abstract}

\begin{IEEEkeywords}
  Robot-Human dialogues, Cooperative principle, Gricean maxims,
  Co-operating Distributed Grammar Systems
\end{IEEEkeywords}

\section{Introduction}
Natural language is one of the easiest and most efficient
means for humans to communicate, and has recently also
been the focus for extensive research in human-robot interaction
(HRI).
A social robot with language capabilities has to understand not only
single utterances but must also be able to conduct a dialogue with a
human.
Human dialogues follow conversational norms in order
to be successful, and phenomena such as sudden changes of
topic, need of clarification, ambiguity, turn taking,
misunderstandings,
and non-understandings influence the character and quality of a
dialogue.
Current approaches to computerised dialogue systems
do not explicitly handle conversational norms.
The overall goal of our research is to conduct work in
this area by formalising dialogue and conversational norms,
and by developing dialogue system components that take
breaches of norms into account.

Our work is divided into the following three parts

\begin{enumerate}
\item Formalising dialogue structure and mental states of dialogue
  participants.
\item Formalising conversational norms occurring in dialogue.
\item Developing computational methods to detect and handle violations
  of conversational norms in dialogue management.
\end{enumerate}

We believe that a formalisation and understanding of how and why
dialogue structure, conversational norms and changes of mental states
co-evolve in the course of utterance exchanges is essential for the
development of computational methods for dialogue management in HRI.

\section{Background}

Dialogues are conversations, intentionally focused to question
thoughts and actions, address problems, increase common knowledge
and hence bring greater understanding~\cite{romney2005art}.
The dialogue structure or dialogue flow is currently not
well understood and existing paradigms to model dialogue
structure fail to generalise or provide insight. The two main
paradigms to dialogue managment are knowledge-based
approaches and data-driven approaches~\cite{Lee2010}. The data-driven
paradigm learns how a dialogue should be conducted from dialogue
corpora,
whereas the knowledge-driven paradigm relies on handcrafted dialogue
flows and thus on expert knowledge.
Data-driven approaches (for example, \cite{Kim16, Thomson2010}, fall short
of providing an understanding into the problem of dialogue
management and can lead to serious ethical consequences\footnote{In
  March 2016,
  Microsoft's chatbot \emph{Tay} parroted racist language after having
  learned from anonymised public data. It was taken offline by
  Microsoft around 16 hours after its launch.}.

The knowledge-based approaches (for example, \cite{hori:2009:wfstbsdm,
  Rama15} are insufficient
in real-world setting as these approaches do not scale
for real applications. Recent hybrid approaches to
dialogue management combine the benefits of both approaches
trying to avoid the disadvantages \cite{Lison2015}. Our approach is a
hybrid approach combining a finite-state and data-driven methods.
Gricean maxims were introduced in \cite{grice1975logic} as a way to describe
how dialogue participants ideally form their utterances
(and thus also what dialogue participants may assume utterances to
be).
Grice views a conversation as a collaborative action where the
participants agree upon a common intention or a predefined
direction. The Gricean maxims are
stated as follows:

\begin{enumerate}
\item Quantity: Make your contribution as informative as
possible.
\item Quality: Do not say what you believe to be false or
which lacks evidence.
\item Relation: Be relevant.
\item Manner: Avoid obscurity and ambiguity. Be brief and
  orderly.
\end{enumerate}

The author in \cite{monz2000modeling} analysed and proposed a model for ambiguous
expressions in multi-agent systems, while in \cite{de2007explaining}
the authors provided a formal model for Grice’s Quantity
implicature for a given utterance.

\section{Approach}

In line with viewing dialogues as collaborative actions, we
formalise dialogues (e.g. turn takes and general dialogue
structure), the mental states of dialouge participants, and conversational
norms with co-operating distributed grammar
systems (CDGSs). CDGSs are abstract devices for describing
multi-agent systems, such as a human and a robot, by means
of formal grammars based on the blackboard architecture
(see, for example, \cite{Csuhaj-Varju:1994:GSG:561869}).
Using CDGS to model dialogue structure allows us to reflect conversational norm as a public string that all agents
(e.g. dialouge participants) work on together, transforming
and extended the string during the dialogue. How the string
is transformed (i.e. how a robot recovers from violations of
conversational norms) is defined by a so-called derivation
mode that the agents are in.
Within our formal framework we investigate how and
why conversational norms are reflected in utterances and
the entire dialogue structure. That is, by formalising conversational
norms we are able to develop computational
methods to identify breaches. For instance, the maxim of brevity (i.e. be brief) can be expressed using the number of
words in a dialogue turn. To express the maxim of relevance,
topic modelling can be used, based on Latent Dirichlet allocation
(LDA) or automated semantic analysis (e.g. analysing
thematic roles). The topic identification is formalised within
our CDGS framework in order to investigate how and why
topics occur during a dialogue (i.e. dialogue structure).
We further develop computational methods to handle
breaches of conversational norms. For example, if a human in
a dialogue is not brief the robot might be allowed to interrupt
the human. After a topic change is identified, the robot can
either follow up the new topic or resume the previous topic
depending on the extent of the violation of the relevance
maxim. If the maxim of informativeness is violated, the robot
switches to a mode in which it either asks for more information
(if the information by the human was too sparse) or
interrupt the human (if the information was too detailed).


\bibliographystyle{IEEEtran}
\bibliography{library.bib}
\end{document}